\documentclass[11pt,a4paper]{article}
\usepackage[hyperref]{acl2019}
\usepackage{times}
\usepackage{latexsym}
\usepackage{booktabs}
\usepackage{color}
\usepackage{graphicx}
\usepackage[TABBOTCAP]{subfigure}
\usepackage[shortlabels]{enumitem}

\usepackage{url}

\aclfinalcopy % Uncomment this line for the final submission
\usepackage{multirow}

\def\CM{{\mathcal C}}

\def\GM{{\mathcal G}}

\def\JM{{\mathcal J}}
\def\KM{{\mathcal K}}

\def\SM{{\mathcal S}}

\def\e{{\bf e}}

\def\k{{\bf k}}
\def\o{{\bf o}}

\def\x{{\bf x}}

\def\0{{\bf 0}}
\def\1{{\bf 1}}

\usepackage{amsmath}

\def\aclpaperid{622} %  Enter the acl Paper ID here

\title{Knowledge-aware Pronoun Coreference Resolution}

\author{
Hongming Zhang$^\clubsuit$\thanks{~This work was partially done during the internship of the first author in Tencent AI Lab.},~ Yan Song$^\spadesuit$,~ Yangqiu Song$^\clubsuit$, and Dong Yu$^\spadesuit$ \\
  $^\clubsuit$Department of CSE, The Hong Kong University of Science and Technology\\
  $^\spadesuit$Tencent AI Lab\\
  { hzhangal@cse.ust.hk, clksong@gmail.com, yqsong@cse.ust.hk, dyu@tencent.com}\\
}

\date{}

\begin{document}
\maketitle
\begin{abstract}
    Resolving pronoun coreference requires knowledge support, especially for particular domains (e.g., medicine).
    In this paper, we explore how to leverage different types of knowledge to better resolve pronoun coreference with a neural model.
    To ensure the generalization ability of our model, we directly incorporate knowledge in the format of triplets, which is the most common format of modern knowledge graphs, instead of encoding it with features or rules as that in conventional approaches.
    Moreover, since not all knowledge is helpful in certain contexts, to selectively use them, we propose a knowledge attention module, which learns to select and use informative knowledge based on contexts, to enhance our model.
    Experimental results on two datasets from different domains prove the validity and effectiveness of our model, where it outperforms state-of-the-art baselines by a large margin.
    Moreover, since our model learns to use external knowledge rather than only fitting the training data, it also demonstrates superior performance to baselines in the cross-domain setting. 
    
\end{abstract}

\vspace{3mm}
\section{Introduction}
% Introduce the importance of the task
Being an important human language phenomenon, coreference brings simplicity for human languages while introducing a huge challenge for machines to process, especially for pronouns, which are hard to be interpreted owing to their weak semantic meanings \cite{ehrlich1981search}.
% The question of how human beings resolve pronouns has long been of interest to both linguistics and natural language processing (NLP) communities, for the reason that pronoun itself has weak semantic meaning~\cite{ehrlich1981search} and brings challenges in natural language understanding. 
As one challenging yet vital subtask of the general coreference resolution, pronoun coreference resolution \cite{hobbs1978resolving} is to find the correct reference for a given pronominal anaphor in the context and has showed its importance in many natural language processing (NLP) tasks, such as machine translation~\cite{mitkov1995anaphora}, dialog systems~\cite{strube2003machine}, information extraction~\cite{edens2003investigation}, and summarization~\cite{steinberger2007two}, etc.
%is the task to determine the correspondence 
%\textcolor{red}{
%To explore solutions for that question, pronoun coreference resolution~\cite{hobbs1978resolving} was proposed.
%As one important yet vital subtask of general coreference resolution, pronoun coreference resolution is to find the correct reference for a given pronominal anaphor in the context and has been shown to be crucial for a series of downstream tasks~\cite{mitkov2014anaphora}, including machine translation~\cite{mitkov1995anaphora}, summarization~\cite{steinberger2007two}, information extraction~\cite{edens2003investigation}, and dialog systems~\cite{strube2003machine}.
%}

% \begin{figure}[t]
%     \centering
%     \includegraphics[width=\linewidth]{image/Example.png}
%     \caption{Demonstration of the pronoun coreference examples, which requires complex knowledge to resolve. Blue bold font refers to the target pronoun,
%     where the correct noun reference and other candidates are marked by green underline and red italic fonts, respectively.}
%     \label{fig:example}
% \end{figure}

\begin{table}[]
    \centering
    \small
    \begin{tabular}{l|p{2.1cm}|p{2.9cm}}
    \toprule
         & Example A & Example B \\
    \midrule
        Sentence & \underline{\textcolor{blue}{The apple}} on the table looks great and I want to eat \textbf{\textcolor{red}{it}}. & Yesterday, the patient took \underline{\textcolor{blue}{the CT scan}} in the hospital and \textbf{\textcolor{red}{it}} showed that she had recovered.\\
        \midrule
         Pronoun & it & it \\
         \midrule
         Answer & The apple & the CT scan \\
         \midrule
        Knowledge & We can eat apples but we cannot eat a table. & A `test' shows results to patients; `the CT scan' is a medical test.\\
        
         \bottomrule
    \end{tabular}
    \caption{Demonstration of two pronoun coreference examples, which require complex knowledge (explained in the table) to resolve. Pronouns and their corresponding mentions
    %that the pronouns refer to
    are marked in bold red and underline blue fonts, respectively.
    % \revisehm{Resolving both examples requires the support of knowledge.}
    }
    \label{tab:example}
\end{table}

In general, to resolve pronoun coreferences, one needs intensive knowledge support.
As shown in Table~\ref{tab:example},
%the resolution of pronoun coreferences typically require the support of knowledge. 
answering the first question requires the knowledge on which object can be eaten (apple v.s. table),
%that we often eat apples but we rarely eat tables
while the second question requires the knowledge that the CT scan is a test (not the hospital) and only tests can show something.
Previously,
rule-based \cite{hobbs1978resolving, nasukawa1994robust, mitkov1998robust, zhang2019aser} and feature-based \cite{ng2005supervised, charniak2009works, li2011pronoun} supervised models were proposed to integrate knowledge to this task.
%
%\textcolor{red}{Conventionally, people design rules \cite{hobbs1978resolving, nasukawa1994robust, mitkov1998robust} or use feature-based supervised models \cite{ng2005supervised, charniak2009works, li2011pronoun} to resolve pronoun coreferences.}
% However, owing to the complexity of knowledge, their performance is limited with no effective way to encode all types of knowledge,
% % knwoeldgewoith restricted knowledge 
% %they can only encode limited types of knowledge and hereby
% especially for such in Table \ref{tab:example}.
However, while easy to incorporate external knowledge, these traditional methods faced the problem of no effective representation learning models can handle such complex knowledge.
%cannot resolve the aforementioned problems.
Later, end-to-end solutions with neural models \cite{DBLP:conf/emnlp/LeeHLZ17,lee2018higher} achieved good performance on the general coreference resolution task.
%\textcolor{red}{
%Until recently, end-to-end solutions~\cite{DBLP:conf/emnlp/LeeHLZ17,lee2018higher} were proposed towards solving the general coreference problem and achieves the state-of-the-art performance.
%}
%However, such solutions
Although such algorithms can effectively incorporate contextual information from large-scale external unlabeled data into the model, they are insufficient to incorporate existing complex knowledge into the representation 
%they can only learn from the training data, which is, in general, insufficient 
for covering all the knowledge one needs to build a successful pronoun coreference system.
In addition, overfitting is always observed on deep models,
whose performance is thus limited in cross-domain scenarios and restricts their usage in real applications \cite{liu-etal-2018-domain,acl-2019-reinforcedTDS}.
Recently, a joint model~\cite{zhang2019-context-pronoun} was proposed to connect the contextual information and human-designed features together for pronoun coreference resolution task (with gold mention support) and achieved the state-of-the-art performance.
However, their model still requires the complex features designed by experts, which is expensive and difficult to acquire, and requires the support of the gold mentions.
% However, such deep models rely heavily on the high quality and large-scale annotated data and often has the problem of overfitting the training data, which limits their generalization ability and usage in real life.

% Two examples are as following:

% \noindent (1) \textit{The dog} is chasing \textit{the cat}, but \textbf{it} climbs the tree.  Which climbs the tree?

%  Answer: \textit{the cat}.
 
% \noindent (2) \textit{The patient} took \textit{the CT scan} in \textit{the hospital} and \textbf{it} shows that she has recovered.  Which shows that the patient has recovered?

%  Answer: \textit{the CT scan}.

% As these knowledge are too complicated to be converted into features, traditional feature-based methods cannot effectively leverage these knowledge.
% On the other hand, current deep models~\cite{DBLP:conf/emnlp/LeeHLZ17,lee2018higher} can only capture the contextual information from the annotated data, which typically cannot cover the aforementioned rich knowledge.

% Knowledge graph (KG) has been used as one of the most popular methods for machines to store and use these knowledge.
% In modern knowledge graphs, the knowledge is often represented as triplets, where two concepts are connected via a typed relation (e.g., (`cat', \textit{can}, `climb the tree'), (`CT scan', \textit{is}, `test')).

To address the limitations of the aforementioned models,
in this paper, we propose a novel end-to-end model that learns to resolve pronoun coreferences with general knowledge graphs (KGs).
Different from conventional approaches, our model does not require to use featurized knowledge.
%to be formatted as features.
Instead, we directly encode knowledge triplets, the most common format of modern knowledge graphs, into our model.
In doing so, the learned model can be easily applied across different knowledge types as well as domains with adopted KG.
%in other domains by changing the used knowledge graphs.
Moreover, to address the knowledge matching issue,
%problem that not all the knowledge are helpful,
we propose a knowledge attention module in our model, which learns to select the most related and helpful knowledge triplets according to different contexts.
Experiments conducted on general (news) and in-domain (medical) cases 
%We conduct experiments with two pronoun resolution coreference datasets:
%CoNLL-2011 shared task (news) and i2b2 (medical),
%The experimental results on both datasets
shows that the proposed model outperforms all baseline models by a great margin.
Additional experiments with the cross-domain setting further illustrate the validity and effectiveness of our model in leveraging knowledge smartly rather than fitting with limited training data\footnote{All code and data are available at: \url{https://github.com/HKUST-KnowComp/Pronoun-Coref-KG}.}.
%In addition, we also conduct experiments with the cross-domain setting to show that our model has better cross-domain performance because our model learns to use knowledge rather than fit the limited training data.
% Our contribution
To summarize, this paper makes the following contributions:
\begin{enumerate}[leftmargin=*]
\setlength{\itemsep}{0pt}
\setlength{\parsep}{0pt}
\setlength{\parskip}{0pt}
    \item We explore how to resolve pronoun coreferences with KGs, which outperforms all existing models by a large margin on datasets from two different domains.
    % \item We explore how to combine deep models and external human knowledge \revisehm{};
    \item We propose a knowledge attention module, which helps to select the most related and helpful knowledge from different KGs.
    \item We evaluate the performance of different pronoun coreference models in a cross-domain setting and show that our model has better generalization ability than state-of-the-art baselines.
\end{enumerate}

% The experiment code is available at: https://github.com/HKUST-KnowComp/Pronoun-coref-kg.

% 

\section{The Task}
\label{sec:task}

%In this section, we formally define the task of end-to-end pronoun coreference resolution.
Given a text $D$, which contains a pronoun $p$, the goal is to identify all the mentions that $p$ refers to.
We denote the correct mentions $p$ refers to as $c \in \CM$, where $\CM$ is the correct mention set.
%of all correct mentions.
Similarly, each candidate span is denoted as $s \in \SM$, where $\SM$ is the set of all candidate spans.
Note that in the case where no golden mentions are annotated, all possible spans in $D$ are used to form $\SM$.
%
%Thus, we denote the set of correct mentions as $\CM$ and the set of all candidate spans $\SM$.
To exploit knowledge,
%Besides that,
we denote the knowledge set as $\GM$, instantiated by multiple knowledge triplets\footnote{Each triplet contains a head, a tail, and a relation from the head to the tail.}.
The task is thus to identify $\CM$ out of $\SM$ with the support of $\GM$.
% which may come from external knowledge and local context.
Formally, it optimizes
%which can be formulated as

\begin{figure}[t]
    \centering
    \includegraphics[width=\linewidth]{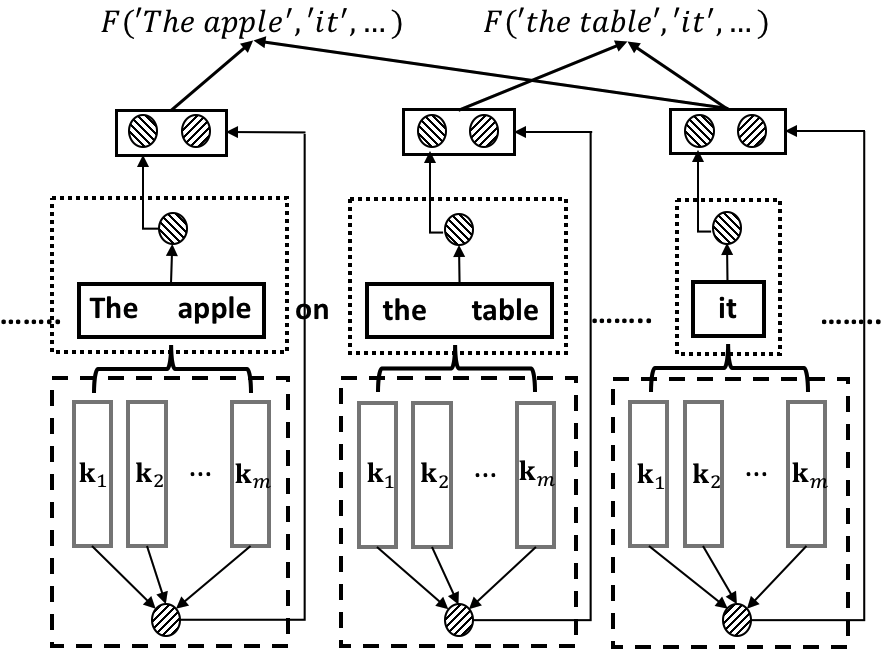}
    \caption{
    The overall framework of our approach to pronoun corference resolution with KGs.
    $\k_1$,...,$\k_m$ represent the retrieved knowledge for each span in the black boxes.
    Dotted box represents the span representation module, which generates a contextual representation for each span.
    Dashed box represents the knowledge selection module, which selects appropriate knowledge based on the context and generates an overall knowledge representation for each span.
    $F(\cdot)$ is the overall coreference scoring function.}
    \label{fig:framework}
    %\vspace{-1em}
\end{figure}

\begin{equation}\label{Objective}
\JM = \frac{\sum_{c \in \CM}{e^{F(c, p, \GM, D)}}}{\sum_{s \in \SM}e^{F(s, p, \GM, D)}},
\end{equation}
where $F(\cdot)$ is the overall scoring function\footnote{We omit $\GM$ and $D$ in the rest of this paper for simplicity.} of $p$ referring to $s$ in $D$ with $\GM$.
%and retrieved knowledge triplets $\TM_s$.
The details of $F$ are illustrated in the following section.
%~\ref{sec:model}.

% Here, each mentions represents a word list that appear in $D$ and we denote the set of correct mentions and all possible mentions as $\CM$ and $\MM$ respectively.

\section{Model}\label{sec:model}

The overall framework of our model is shown in Figure~\ref{fig:framework}.
There are several layers in it.
At the bottom, we encode all mention spans ($s$) and pronouns ($p$) into embeddings so as to incorporate contextual information.
In the middle layer, for each pair of ($s$, $p$), we use their embeddings to select the most helpful knowledge triplets from $\GM$ and generate the knowledge representation of $s$ and $p$.
At the top layer, we concatenate the textual and knowledge representation as the final representation of each $s$ and $p$, and then use this representation to predict whether there exists the coreference relation between them.

\begin{figure}[t]
    \centering
    \includegraphics[width=\linewidth]{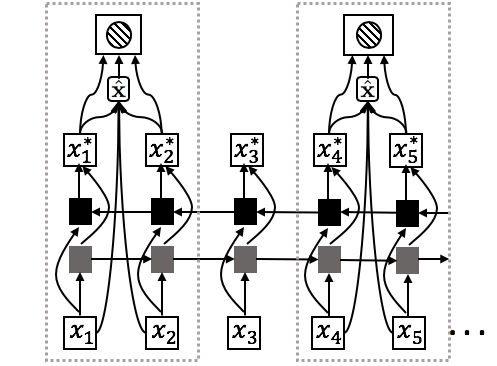}
    \caption{The structure of the span representation module. BiLSTM and attention are employed to encode the contextual information.}
    \label{fig:mention_representation}
    %\vskip -1em
\end{figure}

\subsection{Span Representation}
%\textcolor{red}{[Hongming: this whole section is directly copied from previous paper as the most bottom layer is the same.]}
Contextual information is crucial to distinguish the semantics of a word or phrase,
especially for text representation learning \cite{song-etal-2018-directional,ijcai2018-607}.
In this work, a standard bidirectional LSTM (BiLSTM) \cite{hochreiter1997long} model is used to encode each span with attentions \cite{bahdanau2014neural}, which is similar to the one used in \citet{DBLP:conf/emnlp/LeeHLZ17}.
%Following \citet{DBLP:conf/emnlp/LeeHLZ17}, we adopt the standard bidirectional LSTM (biLSTM) \cite{hochreiter1997long} and the attention mechanism~\cite{bahdanau2014neural} to generate the span representation,
The structure is shown in Figure~\ref{fig:mention_representation}.
Let initial word embeddings in a span $s_i$ be denoted as $\x_1,...,\x_T$
and their encoded representation be $\x^*_1,...,\x^*_T$.
%we denote their representations $\x^*_1,...,\x^*_T$ after encoded by the biLSTM.
%
%\textcolor{red}{
%    \begin{align*}
%    \small
%        f_{t, \delta} &= \sigma(W_{f}[x_t, h_{t+\delta,\delta}] + b_i)\\
%        o_{t, \delta} &= \sigma(W_{o}[x_t, h_{t+\delta,\delta}] + b_o)\\
%        \Tilde{c}_{t, \delta} &= \tanh(W_{c}[x_t, h_{t+\delta,\delta}] + b_c)\\
%        c_{t, \delta} &= f_{t, \delta} \otimes\Tilde{c}_{t, \delta} + (1-f_{t, \delta}) \otimes c_{t+\delta, \delta}\\
%        h_{t, \delta} &= o_{t, \delta} \otimes \tanh( c_{t, \delta})\\
%        x^*_t &= [h_{t, 1}, h_{t, -1}],
%    \end{align*}
%where $\delta \in {-1, 1}$ indicates the directionality of each LSTM,
%}
%Then we obtain
%
The weighted embeddings of each span $\hat{\x_i}$ is obtained by
\begin{equation}\label{eq:overall_embedding}
	\hat{\x_i} = \sum_{t=1}^{T}a_t \cdot \x_t,
\end{equation}
where $a_t$ is the
inner-span attention computed by
% \begin{eqnarray}
% 			\alpha_t = NN_\alpha(x^*_t),\\
% 			a_t = \frac{e^{\alpha_t}}{\sum_{k=START(i)}{END(i)}e^{\alpha_k}},  \label{eq:uprime}
% 		\end{eqnarray}
\begin{equation}\label{eq:mention_attention}
	a_t = \frac{e^{\alpha_t}}{\sum_{k=1}^{T}e^{\alpha_k}},
\end{equation}
where $\alpha_t$ is a standard feed-forward neural network\footnote{We use $NN$ to present feed-forward neural networks.} $\alpha_t$ = $NN_\alpha(\x^*_t)$.

%Using this \revisehm{attention weight} as the weight distribution of different words, we can then get the weighted \revisehm{initial word} embedding with:

%
%Afterwards, we concatenate
Finally,
the starting ($\x^*_{start}$) and ending ($\x^*_{end}$) embedding of each span is concatenated with the weighted embedding ($\hat{\x_i}$) and the length feature ($\phi(i)$)
to form its final representation $\e$:
\begin{equation}\label{eq:mention_embedding}
    \e_i = [\x^*_{start},\x^*_{end},\hat{\x_i},\phi(i)].
\end{equation}
Thus the span representation of $s$ and $p$ are marked as $\e_s$ and $\e_p$, respectively.

\begin{figure}[t]
    \centering
    \includegraphics[width=\linewidth]{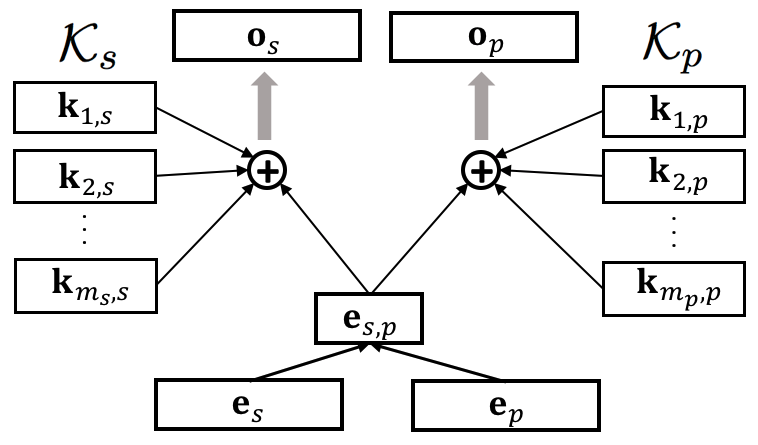}
    \caption{The structure of the knowledge attention module. The joint representation of the candidate span and pronoun is used to select knowledge for $s$ and $p$.}
    \label{fig:kg_attention}
\end{figure}

%\subsection{Knowledge Retrieval}

\subsection{Knowledge Representation}

For each candidate span $s$ and the target pronoun $p$, different knowledge from a KG can be extracted with various methods.
%but we simply use string match for simplicity.
For simplicity and generalization consideration, we use the string match in our model for knowledge extraction.
Specifically, for each triplet $t \in \GM$ where the head and tail of $t$ are both lists of words, if its head is the same as the string of $s$, we consider it to be a related triplet.
Therefore, we encode the information of $t$ 
%the information of \textcolor{red}{$t$'s tail}
by the averaging embeddings of all words in its tail.
For example, if $s$ is `the apple' and the knowledge triplet (`the apple', \textit{IsA}, `healthy food') is found by searching the KG, we represent this relation from the averaged embeddings of `healthy' and `food'.
Consequently, for $s$ and $p$, we denote their retrieved knowledge set as $\KM_s$ and $\KM_p$ respectively, where $\KM_s$ contains $m_s$ related knowledge embeddings $\k_{1,s}$, $\k_{2,s}$, ..., $\k_{m_s, s}$ and $\KM_p$ contains $m_p$ of them $\k_{1,p}$, $\k_{2,p}$, ..., $\k_{m_p, p}$.
% Therefore, for each $s$, we denote the retrieved knowledge set as $\KM_s$, which contains $m_s$ related knowledge embeddings \textcolor{red}{$k_1$, $k_2$, ..., $k_{m_s}$}.

To incorporate the aforementioned knowledge embeddings into our model,
we face a challenge that there are a huge number of such embeddings while most of them are useless in certain contexts.
%challenge we are facing of injecting knowledge into the model is that the number of knowledge could be huge and most of them are useless in one specific case.
%Injecting all of them into the model could bring huge computational burden and unnecessary noise.
%To address this obstacle,
%For this challenge,
To solve it,
a knowledge attention module is proposed to select the appropriate knowledge.

For each pair of ($s$, $p$), 
% assuming the retrieved knowledge set of $s$ and $p$ denoted by $\KM_s$ and $\KM_p$ respectively,
as shown in Figure~\ref{fig:kg_attention}, we first concatenate $\e_s$ and $\e_p$ to get the overall (span, pronoun) representation $\e_{s,p}$, which is used to select knowledge for both $s$ and $p$.
Taking that for $s$ as example,
we compute the weight of each $\k_i \in \KM_s$ by
%After that, for each $\k_i \in \KM_s$, we can compute its weight by
\begin{equation}
    w_i = \frac{e^{\beta_{\k_i}}}{\sum_{\k_j \in \KM_s}{e^{\beta_{\k_j}}}},
\end{equation}
where $\beta_{\k} = NN_\beta([\e_{s,p},\k])$.
As a result, the knowledge of $s$ is summed by
\begin{equation}
    \o_s = \sum_{\k_i \in \KM_s}w_i \cdot \k_i.
\end{equation}
%Similarly, we can get $\o_p$, which is
to represent
the overall knowledge for $s$.
A similar process is also conducted for $p$ with its knowledge representation $\o_p$.

\subsection{Scoring}
%For $s$ and $p$, we then compute the overall coreference score function as follows:
The final score of each pair ($s$, $p$) is computed by
\begin{equation}
    F(s,p) = f_m(s) + f_c(s,p),
\end{equation}
where $f_m(s) = NN_m([\e_s,\o_s])$ is the scoring function for $s$ to be a valid mention and $f_c(s,p) = NN_c([\e_n, \o_n, \e_p, \o_p, \e_n \odot \e_p, \o_n \odot \o_p])$ is the scoring function to identify whether there exists a coreference relation from $p$ to $s$, with $\odot$ denoting element-wise multiplication.

%\subsection{Softmax Selection}
After getting the coreference score for all mention spans, we adopt a softmax selection on the most confident candidates for the final prediction,
which is formulated as
%Formally, the module takes $F(s,p)$ as input for each $s$, uses a softmax computation:
%Specifically, we compute the softmax for each $n$ by
\begin{equation}\label{eq:softmax_pruning}
        \hat{F}(s, p) = \frac{e^{F(s, p)}}{\sum_{s_i \in \SM}e^{F(s_i, p)}}.
\end{equation}
where candidates with score $\hat{F}$ higher than a threshold $t$ are selected.

\section{Experiments}
%In this section, we introduce the experimental details.
Experiments are illustrated in this section.

\subsection{Datasets}
%  \begin{table}[t]
%     \centering
%     \small
%     \subtable[CoNLL]{
%       \begin{tabular}{l|ccc|c}
%         \toprule
%         Type            & train  & dev   & test  & all    \\
%          \midrule
%          Third Personal & 21,828 & 2,518 & 3,530 & 27,876 \\
%          Possessive     & 7,749  & 1,007 & 1,037 & 9,793  \\
%         \midrule
%         All     & 29,577 & 3,525 & 4,567 & 37,669 \\
%         \bottomrule
%     \end{tabular}
%     }
%     \subtable[i2b2]{
%       \begin{tabular}{l|cc|c}
%         \toprule
%         Type            & train    & test  & all    \\
%          \midrule
%          Third Personal & 2,024  & 1,244 & 3,268 \\
%          Possessive     & 685   & 367 & 1,052  \\
%         \midrule
%         All     & 2,709  & 1,611 & 4,320 \\
%         \bottomrule
%     \end{tabular}
%     }
%     \caption{Statistics of the evaluation datasets. Number of selected pronouns are reported.}
%     \label{tab:statics}
% \end{table}

Two datasets are used in our experiments, where they are from two different domains:
%In this subsection, we introduced the used dataset. 
%To test the effectiveness of different pronoun coreference resolution models, we select two representative datasets from two different domains.
%One is from the news domain and the other one is from the medical domain. The details about these datasets are as follows:
\begin{itemize}[leftmargin=*]
\setlength{\itemsep}{0pt}
\setlength{\parsep}{0pt}
\setlength{\parskip}{0pt}
    \item \textbf{CoNLL:} The CoNLL-2012 shared task~\cite{pradhan2012conll} corpus,
    %is used as the evaluation dataset,
    which is a widely used dataset selected from the Ontonotes 5.0\footnote{https://catalog.ldc.upenn.edu/LDC2013T19}.
    \item \textbf{i2b2:} The i2b2 shared task dataset~\cite{uzuner2012evaluating}, consisting of electronic medical records from two different organizations, namely, Partners HealthCare (Part) and Beth Israel Deaconess medical center (Beth).
    All records have been fully de-identified and manually annotated with coreferences.
\end{itemize}

We split the datasets into different proportions based on their original settings.
%, with the statistics reported in Table \ref{tab:statics}.
%We consider
Three types of pronouns are considered in this paper following \citet{ng2005supervised}, i.e., third personal pronoun (e.g., \textit{she}, \textit{her}, \textit{he}, \textit{him}, \textit{them}, \textit{they}, \textit{it}), possessive pronoun (e.g., \textit{his}, \textit{hers}, \textit{its}, \textit{their}, \textit{theirs}), and demonstrative pronoun (e.g., \textit{this}, \textit{that}, \textit{these}, \textit{those}).
Table \ref{tab:statics} reports the number of the three types of pronouns and the overall statistics of the experiment datasets with proportion splittings.
Following conventional approaches \cite{ng2005supervised, li2011pronoun}, for each pronoun,
%in a text,
we consider its candidate mentions from the previous two sentences and the current sentence it belongs to.
According to our selection range of the candidate mentions,
%on average,
each pronoun in the CoNLL data and i2b2 data has averagely 1.3 and 1.4 correct references, respectively.

\begin{table}[t]
    \centering
    \small
    \begin{tabular}{c|l|rrr|r}
    \toprule
        \multicolumn{2}{c|}{Dataset} & TP & Poss & Dem & All \\
        \midrule
         \multirow{ 3}{*}{CoNLL} & train & 21,828 & 7,749 & 2,229 & 31,806 \\
                                & dev & 2,518 & 1,007 & 222 & 3,747 \\
                                & test & 2720 & 1,037 & 321 &  4,078 \\
        \midrule
        \multirow{ 2}{*}{i2b2} & train & 2,024 & 685 & 270 & 2,979 \\
                                & test & 1,244 & 367 & 166 & 1,777 \\
        \midrule
        \multicolumn{2}{c|}{Overall} & 30,334 & 10,845 & 3,208 & 44,387 \\
        \bottomrule
    \end{tabular}
    \caption{Statistics of the two datasets.
    `TP', `Poss', and `Dem' refer to third personal, possessive, and demonstrative pronouns, respectively.}
    \label{tab:statics}
\end{table}

\subsection{Knowledge Resources}

As mentioned in previous sections, our model is designed to leverage general KGs, where it takes triplets as the input of knowledge representations.
%As introduced in Section~\ref{sec:task}, the proposed model is designed for the general knowledge graphs, where knowledge are stored in the format of triplets.
For all knowledge resources, we format them as triplets and 
%Thus we firstly formalize all the knowledge into the format of triplets and then
merge them together to obtain the final knowledge set.
%The details about the used knowledge resources are as follows:
Different knowledge resources are introduced as follows.

\vspace{0.1cm}
\noindent\textbf{Commonsense knowledge graph (OMCS).}
We use the largest commonsense knowledge base, the open mind common sense (OMCS) \cite{singh2002open} in this paper.
OMCS contains 600K crowd-sourced commonsense triplets such as \textit{(food, UsedFor, eat)} and \textit{(wind, CapableOf, blow to east)}.
All relations in OMCS are human-defined and we select those highly-confident ones (confidence score larger than $2$) to form the OMCS KG, with 62,730 triplets.

\vspace{0.1cm}
\noindent \textbf{Medical concepts (Medical-KG).}
Being part of the i2b2 contest, the related knowledge about medical concepts such as \textit{(the CT scan, is, test)} and \textit{(intravenous fluids, is, treatment)} are provided. 
The annotated triplets are used as the medical concept KG, which contains 22,234 triplets.

\vspace{0.1cm}
\noindent \textbf{Linguist features (Ling).}
In addition to manually annotated KGs, we also consider linguist features, i.e., plurality and animacy \& gender (AG), as one important knowledge resources.
Stanford parser\footnote{https://stanfordnlp.github.io/CoreNLP/} is employed to generate plurality, animacy, and gender markups for all the noun phrases, so as to automatically generate linguistic knowledge (in the form of triplets) for our data.
Specifically, the plurality feature denotes each $s$ and $p$ to be singular or plural.
The animacy \& gender (AG) feature denotes whether the $n$ or $p$ is a living object, and being male, female, or neutral if it is alive.
For example, 
a mention `the girls' is labeled as \textit{plural} and \textit{female}; we use triplets \textit{(`the girls', plurality, Plural)} and \textit{(`the girls', AG, female)} to represent them.
%into the linguist feature KG.
As a result, we have 40,149 and 40,462 triplets for plurality and AG, respectively.

\vspace{0.1cm}
\noindent\textbf{Selectional Preference (SP).}
Selectional preference \cite{hobbs1978resolving} knowledge is employed as the last knowledge resource, which is the semantic constraint for word usage.
SP generally refers to that, given a predicate (e.g., verb), people have the preference for the argument (e.g., its object or subject) connected.
To collect SP knowledge, we first parse the English Wikipedia\footnote{https://dumps.wikimedia.org/enwiki/} with the Stanford parser and extract all dependency edges in the format of \textit{(predicate, argument, relation, number)}, where predicate is the governor and argument the dependent in each dependency edge\footnote{In the Stanford parser, an `nsubj' edge is created between its predictive and subject when a verb is a linking verb (e.g., am, is); the predicative is thus treated as the predicate for the subject (argument) in this paper.}.
Following \cite{resnik1997selectional},
each potential SP pair is measured by a posterior probability
%as the SP strength measure.
%The posterior probability is defined as follows:
\begin{equation}\label{naturalprobability}
            P_r(a|p) = \frac{Count_r(p, a)}{Count_r(p)}, 
\end{equation}
where $Count_r(p)$ and $Count_r(p, a)$ refer to how many times $p$ and the predicate-argument pair ($p$, $a$) appear in the relation $r$, respectively.
In our experiment, if $P_r(a|p) > 0.1 $ and $Count_r(p,a) > 10$, we consider the triplet ($p$, $r$, $a$) (e.g., \textit{(`dog', nsubj, `barks')}) a valid SP relation.
Finally, we select two SP relations, \textit{nsubj} and \textit{dobj}, to form the SP knowledge graph, including 
%As the result, we collect
17,074 and 4,536 frequent predicate-argument pairs for \textit{nsubj} and \textit{dobj}, respectively.

\begin{table*}[t]
\small
    \centering
    \subtable[CoNLL]{
    \begin{tabular}{l|ccc|ccc|ccc|ccc}
        \toprule
         \multirow{ 2}{*}{Model} & \multicolumn{3}{c|}{Third Personal} & \multicolumn{3}{|c|}{Possessive} & \multicolumn{3}{c|}{Demonstrative} &  \multicolumn{3}{|c}{All}\\
         & P & R & F1 & P & R & F1 & P & R & F1 & P & R & F1 \\
         
         \midrule
         Deterministic & 25.5 & 58.9 & 35.6 & 22.9 & 64.3 & 33.8 & 3.4 & 5.7 & 4.2 & 23.4 & 57.0 & 33.4  \\ 
         Statistical & 25.8 & 62.1 & 36.5 & 28.9 & 64.9 & 40.0 & 9.8 & 6.3 & 7.6 & 25.4 & 59.3 & 36.5 \\ 
         Deep-RL & 78.6 & 63.9 & 70.5 & 73.3 & 68.9 & 71.0 & 3.7 & 2.9 & 5.5 & 76.4 & 61.2 & 68.0 \\ 
         \midrule
         End2end & 70.6 & 75.7 & 73.1 & 73.0 & 76.2 & 74.6 & 58.4 & 17.6 & 27.0 & 71.1 & 72.1 & 71.6 \\
         \midrule 
         Without KG & 78.2 & 72.4 & 75.2 & 80.0 & 66.4 & 72.6 & 46.7 & 62.5 & 53.4 & 75.7 & 70.1 & 72.8 \\
         Without Attention & 76.6 & 77.9 & 77.2 & 79.0 & 73.5 & 76.2 & 42.4 & 72.6 & 53.5 & 73.6 & 76.4 & 74.9 \\
         Our Complete Model & 78.8 & 77.8 & \textbf{78.1} & 80.7 & 72.5 & \textbf{76.4} & 45.3 & 66.7 & \textbf{53.9} & 75.9 & 75.6 & \textbf{75.7} \\
         \bottomrule
    \end{tabular}
    }
    \subtable[i2b2]{
    \begin{tabular}{l|ccc|ccc|ccc|ccc}
        \toprule
         \multirow{ 2}{*}{Model} & \multicolumn{3}{c|}{Third Personal} & \multicolumn{3}{|c|}{Possessive} & \multicolumn{3}{c|}{Demonstrative} &  \multicolumn{3}{|c}{All}\\
         & P & R & F1 & P & R & F1 & P & R & F1 & P & R & F1 \\
         \midrule
         Deterministic & 25.7 & 57.4 & 35.5 & 25.2 & 61.6 & 35.7 & 6.6 & 4.0 & 5.0 & 25.1 & 54.0 & 34.3  \\ 
         Statistical & 19.3 & 35.9 & 25.1 & 25.7 & 50.5 & 34.0 & 6.7 & 4.5 & 5.4 & 20.5 & 36.6 & 26.3 \\ 
         Deep-RL & 78.2 & 48.0 & 59.5 & 78.6 & 57.7 & 66.5 & 9.1 & 5.1 & 9.6 & 77.8 & 46.3 & 58.1 \\
         \midrule
         End2end & 95.0 & 93.4 & 94.2 & 95.3 & 96.0 & 95.7 & 74.8 & 52.5 & 61.7 & 93.9 & 90.7 & 92.3 \\
         \midrule 
         Without KG & 96.8 & 95.9 & 96.3 & 97.1 & 97.5 & 97.3 & 66.5 & 68.2 & 67.3 & 94.3 & 94.0 & 94.2 \\
         Without Attention & 96.1 & 97.2 & 96.6 & 96.3 & 98.2 & 97.2 & 66.7 & 77.8 & 71.8 & 93.4 & 95.9 & 94.6 \\
         Our Complete Model & 97.5 & 96.3 & \textbf{96.9} & 98.5 & 97.8 & \textbf{98.2} & 71.9 & 72.2 & \textbf{72.0} & 95.6 & 94.7 & \textbf{95.2} \\
         \bottomrule
    \end{tabular}
    }
    \caption{The performance of pronoun coreference resolution with different models on two evaluation datasets. Precision (P), recall (R), and the F1 score are reported, with the best one in each F1 column marked as bold.
    %and we use F1 as the main evaluation matrix.
    }
    \label{tab:main_result}
\end{table*}

\subsection{Baselines}

Several baselines are compared in this paper, including 
%We firstly compare with
three widely used pre-trained models:
\begin{itemize}[leftmargin=*]
\setlength{\itemsep}{0pt}
\setlength{\parsep}{0pt}
\setlength{\parskip}{0pt}
\item
\textbf{Deterministic} model \cite{raghunathan2010multi}, which is an unsupervised model and leverages manual rules to detect coreferences.
\item
\textbf{Statistical} model \cite{clark2015entity}, which is a supervised model and trained on manually crafted entity-level features between clusters and mentions.
\item
\textbf{Deep-RL} model \cite{DBLP:conf/emnlp/ClarkM16}, which uses reinforcement learning to directly optimize the coreference matrix instead of the loss function of supervised learning.
\end{itemize}
%Other than these conventional methods, we also compare with one deep model, which is the current state-of-the-art model.
The above models are included in the Stanford CoreNLP toolkit\footnote{https://stanfordnlp.github.io/CoreNLP/coref.html}.
We also include a state-of-the-art end-to-end neural model as one of our baselines:
\begin{itemize}[leftmargin=*]
\item
\textbf{End2end} \cite{lee2018higher}, which is the current state-of-the-art model performing in an end-to-end manner and leverages both contextual information and a pre-trained language model \cite{peters2018deep}.
\end{itemize}
%Note that the Deterministic, Statistical, and Deep-RL models are included in the Stanford CoreNLP toolkit\footnote{https://stanfordnlp.github.io/CoreNLP/coref.html}, experiments are conducted with their provided model.
%For End2end, we
We use their released code\footnote{https://github.com/kentonl/e2e-coref}.
In addition,
%Last but not least,
to show the importance of incorporating knowledge,
%and the effectiveness of the proposed model,
we also experiment with two variations of our model:
\begin{itemize}[leftmargin=*]
\setlength{\itemsep}{0pt}
\setlength{\parsep}{0pt}
\setlength{\parskip}{0pt}
\item
\textbf{Without KG} removes the KG component and keeps all other components in the same setting as that in our complete model.
\item
\textbf{Without Attention} removes the knowledge attention module and concatenates all the knowledge embeddings. All other components are identical as our complete model.
\end{itemize}

\subsection{Implementation}

Following the previous work \cite{lee2018higher}, we use the concatenation of the 300d GloVe embeddings~\cite{pennington2014glove} and the ELMo~\cite{peters2018deep} embeddings as the initial word representations for computing span representations.
For knowledge triplets, we use the
%\revisehm{We encode knowledge triplets by the average 
GloVe embeddings to encode tail words in them.
Out-of-vocabulary words are initialized with zero vectors.
% Hyper-parameters are set as follows.
The hidden state of the LSTM module is set to 200, and all the feed-forward networks have two 150-dimension hidden layers.
% The default selected knowledge limitation is set to 5 and selection threshold is set to $10^{-5}$.
The selection thresholds are set to $10^{-2}$ and $10^{-8}$ for the CoNLL and i2b2 dataset, respectively.

For model training, we use cross-entropy as the loss function and Adam \cite{kingma2014adam} as the optimizer.
All the aforementioned hyper-parameters are initialized randomly, and we apply dropout rate 0.2 to all hidden layers in the model.
% Our model treats a candidate as the correct reference if its predicted overall score $F(n,p)$
% is larger than 0.
For the CoNLL dataset, the model training is performed with up to 100 epochs, and the best one is selected based on its performance on the development set. 
For the i2b2 dataset, because no dev set is provided, we train the model up to 100 epochs and use the final converged one.

%\section{Experimental Results}

\subsection{Results}

% \begin{table*}[t]
% \small
%     \centering
%     \begin{tabular}{l|ccc|ccc|ccc|ccc|ccc|ccc}
%         \toprule
%          \multirow{ 2}{*}{Model} & \multicolumn{3}{c|}{Third Personal} & \multicolumn{3}{|c|}{Possessive} &  \multicolumn{3}{|c}{All}\\
%          & P & R & F1 & P & R & F1 & P & R & F1 \\
         
%          \midrule

%          Recent Candidate & 50.7 & 40.0 & 44.7 & 64.1 & 45.5 & 53.2 & 54.4 & 41.6 & 47.2 \\ 
%          Deterministic & 68.7 & 59.4 & 63.7 & 51.8 & 64.8 & 57.6 & 62.3 & 61.0 & 61.7 & 68.7 & xxx & xxx & xxx & xxx & xxx & xxx & xxx & xxx \\ 
%          \midrule

%          Statistical & 69.1 & 62.6 & 65.7  & 58.0 & 65.3 & 61.5  & 65.3 & 63.4 & 64.3 \\ 
%          Deep-RL & 72.1 & 68.5 & 70.3  & 62.9 & 74.5 & 68.2 & 68.9 & 70.3 & 69.6 \\ 
%          End2end & 75.1 & 83.7 & 79.2 & 73.9 & 82.1 & 77.8 & 74.8 & 83.2 & 78.8 \\ 
         
%          \midrule 
%         %  Our Base Model & 73.0 & 89.5 & 80.4  & 73.0 & 88.3 & 80.0 & 73.0 & 89.2 & 80.3 \\
%          KG-embedding & xxx & xxx & xxx & xxx & xxx & xxx & xxx & xxx & xxx \\
%          Our Model & xxx & xxx & xxx & xxx & xxx & xxx & xxx & xxx & xxx \\
%          \bottomrule
%     \end{tabular}
%     \caption{Pronoun coreference resolution performance of different models on the evaluation dataset. Precision (P), recall (R), and F1 score are reported, with the best one in each F1 column marked bold.
%     %and we use F1 as the main evaluation matrix.
%     }
%     \label{tab:main_result}
% \end{table*}

Table~\ref{tab:main_result} reports the performance of all models, with the results for CoNLL and i2b2 in (a) and (b), respectively.
%compares the performance of our model with all baselines.
Overall, our model outperforms all baselines on two datasets 
%performs the best
with respect to all pronoun types.
There are several interesting observations.
%
%We first have some interesting findings about the difference between different datasets and pronoun types.
In general, the i2b2 dataset seems simpler than the CoNLL dataset, which might because that i2b2 only involves clinical narratives and its training data is highly similar to the test data.
As a result,
%And that is why all the deep models
all neural models
perform dramatically good, especially on the third personal and possessive pronouns.
In addition, we also notice that it is more challenging for all models to resolve demonstrative pronouns (e.g., \textit{this}, \textit{that}) on both datasets, because such pronouns may refer to complex things and occur with low frequency.
%\revisehm{may refer to more complex things and appear quite rare}, it is more challenging for all models.
%to resolve demonstrative pronouns well on both datasets.
% , while ours outperform others with a huge margin.
% While for 
% %On the other hand, as
% demonstrative pronouns (e.g., \textit{this}, \textit{that}) referring to many different things (e.g., test, treatment), which 
% the performance of \textcolor{red}{all models} \textcolor{red}{what is the ``all model''?} on demonstrative pronouns is still not satisfying.
%
%Besides that, we also have some interesting findings about different models.
%Second,

\begin{table}[t]
    \centering
    \small
    \begin{tabular}{l|cc|cc}
        \toprule
         & \multicolumn{2}{c|}{CoNLL} & \multicolumn{2}{c}{i2b2}\\
         & F1 &$ \Delta$ F1 & F1 &$ \Delta$ F1 \\
        \midrule
        % Our model (ensemble) & 79.5 & 90.2 & 84.5 & +0.7 \\
        The Complete Model  & 75.7 & - & 95.2 & - \\
        \midrule
        ~--OMCS & 74.8 & -0.9 & 95.1 & -0.1 \\
        ~--Medical-KG & 74.5 & -1.2 & 94.6 & -0.6 \\
        ~--Ling & 73.8 & -1.9 & 94.9 & -0.3 \\
        ~--SP & 74.0 & -1.7 & 94.7 & -0.5 \\
        \bottomrule
    \end{tabular}
    \caption{The performance of our model with removing different knowledge resources.
    The F1 of each case and the difference of F1 between each case and the complete model are reported.
    %All of the knowledge resources contribute to the final success of our model but the most important knowledge resources for different datasets are different (Ling for CoNLL and Medical-KG for i2b2).
    % the knowledge graph attention module contributes to the similar performance gap between our model and the KG embedding one.
    }
    \label{tab:abalation}
\end{table}

Moreover, there are significant gaps in the performance of different models, with the following observations.
First, models with manually defined rules or features, which cannot cover rich contextual information, perform poorly.
In contrast,
deep learning models (e.g., End2end and our proposed models), which leverage text representations for context, outperform other approaches by a great margin, especially on the recall.
Second, adding knowledge in an appropriate manner within neural models is helpful, which is supported by that
our model outperforms the End2end model and the Without KG one on both datasets, especially CoNLL, where the external knowledge plays a more important role.
%
%Third, as the original end2end model employs a cluster-based prediction method, where each pronoun is linked to the most possible antecedent it refers to. 
%This strategy makes sure the overall precision of referring pronoun to noun phrases (NPs), but the overall recall can be sacrificed.
%Besides that, we also notice that End2end often group same pronouns across sentences into the same cluster, even though they are refer to different noun phrases, which simply because their representations are very similar with each other.
Third, the knowledge attention module ensures our model to predict more precisely, which also results in the overall improvement on F1.
To summarize, the results suggest that external knowledge is important for effectively resolving pronoun coreference, where rich contextual information determines the appropriate knowledge with a well-designed module.

\section{Analysis}

Further analysis is conducted in this section regarding the effect of different knowledge resources, model components, and settings.
%for hyper-parameters, and the effect of gold mentions.
Details are illustrated as follows.
%in the following subsections.

\subsection{Ablation Study}
We ablate different knowledge for their contributions in our model, with the results reported in Table \ref{tab:abalation}.
%
%\textcolor{red}{To illustrate the importance of different knowledge sources and the knowledge attention mechanism, we ablate various components of our model and report the corresponding F1 scores on the test data.
%The results are shown in Table~\ref{tab:abalation}, which clearly show the necessity of the knowledge.}
%
It is observed that all knowledge resources contribute to the final success of our model,
where different knowledge types play their unique roles in different datasets.
For example, the Ling knowledge contributes the most to the CoNLL dataset while the medical knowledge is the most important one for the medical data. 
% \reviseyq{(pay attention to the modification of prepositional words.)}
%
% In addition to knowledge types, the results also prove the effectiveness of the knowledge attention module,
% which \textcolor{red}{contributes to the performance gap between our model and the KG-embedding one.}
% Without this module, extra noise weakens the effect of knowledge incorporation.

%
\begin{figure}[t]
    \centering
    \includegraphics[width=\linewidth, trim=0 0 0 30]{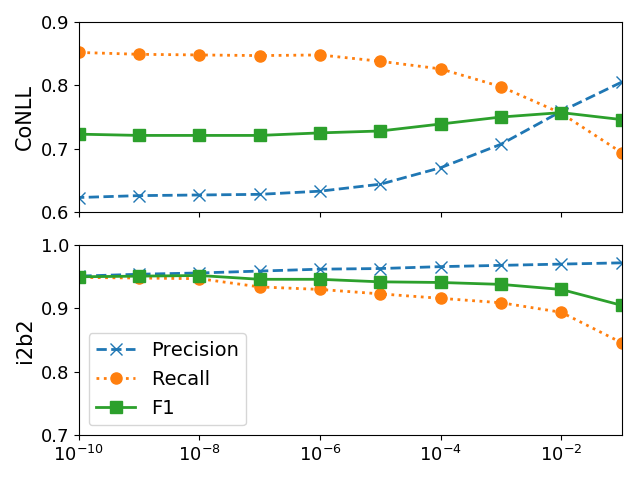}
    \caption{Effect of different softmax selection thresholds with respect to our model performance on two datasets.
    %\revisehm{Influence of different pruning thresholds on the final performance of our model.
    In general, with the threshold becoming larger, less candidates are selected, the precision thus increases while the recall drops.}
    \label{fig:selection}
\end{figure}

\subsection{Effect of the Selection Threshold}
%Hyper-parameter Analysis

We experiment with different thresholds $t$ for the softmax selection.
The effects of $t$ against overall performance are shown in Figure~\ref{fig:selection}.
In general, with the increase of $t$, fewer candidates are selected.
Therefore, the overall precision increases and the recall drops.
Consider that both the precision and recall are important for resolving pronoun coreference, we select different thresholds for different datasets to ensure the balance between precision and recall.
In detail, for the CoNLL dataset,
we set $r=10^{-2}$ to select the most confident predictions;
and for the i2b2 dataset,
we set $r=10^{-8}$ so as to keep more predictions.

\subsection{Effect of Gold Mentions}

\begin{table}[t]
\small
    \centering
    \begin{tabular}{l|c|cc}
        \toprule
        
         Model & Setting & CoNLL & i2b2\\
        
        \midrule
         \multirow{ 2}{*}{End2end} & Original & 71.6  & 92.3  \\
                                  & + Gold mention & 77.8  & 94.4  \\
                                  \midrule
        \multirow{ 2}{*}{Our Model} & Original & 75.7 & 95.2  \\
                                    & + Gold mention & \textbf{80.7} & \textbf{96.0} \\
                                    \midrule
        \cite{zhang2019-context-pronoun} & + Gold mention & 79.9 & -\\

        \bottomrule
    \end{tabular}
    \caption{Influence of gold mentions. F1 scores on different test sets are reported.
    Adding human-annotated gold mentions help both the End2end and our model. Best performed model are indicated with the bold font.
    }
    \label{tab:gold_mention}
\end{table}

The effect of adding gold mentions is shown in Table~\ref{tab:gold_mention}.
Providing gold mentions to the End2end model can significantly boost its performance by 6.2 F1 and 2.1 F1 on the CoNLL and i2b2 dataset, respectively.
Yet, the performance gain from gold mentions is less for our model.
Such results clearly illustrate that our model is able to benefit the mention detection with the help of KG incorporation.
Besides that, with the help of gold mentions, our model achieves the comparable (slightly better) performance with the context-and-knowledge model~\cite{zhang2019-context-pronoun}.
% , which indicates that the proposed model can better incorporate external knowledge due to its generalization ability\footnote{More knowledge types can be supported by our model.}.
As their features are originally designed for CoNLL, we only report the performance on CoNLL in Table~\ref{tab:gold_mention}. 
As we also included one new challenging pronoun type, the demonstrative pronoun, the overall performance of their model is lower than the one reported in the original paper.
The reason of our model being better is that more knowledge resources (e.g., OMCS) can be incorporated into our model due to its generalizable design.
Moreover, it is more difficult for their method~\cite{zhang2019-context-pronoun} to incorporate mention detection into the model, because in this case we need to enumerate all mention spans and generate corresponding features for all spans. which is expensive and difficult to acquire.

% Such results clearly illustrate the capability of our model in effectively performing on less confident candidates to enhance pronoun coreference resolution,
% with the help of KG incorporation.

\subsection{Cross-domain Evaluation}

\begin{table}[t]
\small
    \centering
    \begin{tabular}{l|c|cc}
        \toprule
      \multirow{ 2}{*}{Model} & \multirow{ 2}{*}{Training data} & \multicolumn{2}{c}{Test data}\\
        && CoNLL & i2b2\\
        \midrule
         \multirow{ 2}{*}{End2end} & CoNLL & 71.6  & 75.2  \\
                                  &i2b2 & 20.0  & 92.3  \\
                                  \midrule
        \multirow{ 2}{*}{Our Model} &CoNLL & 75.7 & 80.9  \\
                                    &i2b2 & 42.7 & 95.2 \\

        \bottomrule
    \end{tabular}
    % \begin{tabular}{cc|c|c}
    %     \toprule
    %   Training data & Test data & End2end & Our model\\
    %     \midrule
    %      CoNLL & i2b2 & 75.2 & xxx\\
    %      \midrule
    %      i2b2 & CoNLL & 20.0 & xxx\\

    %     \bottomrule
    % \end{tabular}
    \caption{Cross-domain performance of different models. F1 on the target domain test sets are reported.}
    \label{tab:cross-domain}
    \vspace{-0.2in}
\end{table}

Considering that neural models are intensive data-driven and normally restricted by data nature,
they are not easily applied in a cross-domain setting.
However, if a model is required to perform in real applications, it has to show promising performance on those cases out of the training data.
%One criticize that deep learning models often received is that they are only fitting the datasets and can not be easily applied in another domain.
Herein we investigate 
%In this section, we firstly investigate 
the performance of different models with training and testing in different domains,
%under the cross-domain setting. 
%In detail, we train a model with the dataset in one domain and directly apply the learned model on the other dataset, which is from the totally different domain.
with the results reported
%As shown
in Table~\ref{tab:cross-domain}.
Overall, all models perform significantly worse if they are used cross domains.
Specifically, if we train the End2end model on the CoNLL dataset and test it on the i2b2 dataset, it only achieves 75.2 F1.
As a comparison, our model can achieve 80.9 F1 in the same case.
This observation confirms the capability of knowledge where our model is able to handle.
%The reason behind is that our model learns to use knowledge rather than fitting the training data.
%Similarly, if we train the models with the i2b2 dataset and test them on the CoNLL dataset, End2end and our model can achieve 20.0 and 42.7 F1 respectively.
A similar observation is also drawn for the reversed case.
However, even though our model outperforms the End2end model by 22.7 F1 from i2b2 to CoNLL, its overall performance is still poor, which might be explained by that the i2b2 is an in-domain dataset and the knowledge contained in its training data is rarely useful for the general (news) domain dataset.
Nevertheless, this experiment clearly shows that the generalization ability of deep models is still crucial for building a successful coreference model,
% , which may result in good performance when training data is appropriate.
% However, if the training data does not satisfy the test scenario, 
and learns to use knowledge is a promising solution to it.
%Similarly, the performance of our model also drops by xxx F1 and XXX F1 on the CoNLL and i2b2 datasets respectively.
% Compared to the End2end model, ours narrowed the performance gap by 5\% (51.6 to 47.8) for the CoNLL data and 45\% (18.4 to 10.2) for the i2b2 data, which can be explained by the fact that our model incorporates different knowledge instead of only fitting the training data.

% To further analyze how different parts of our model contributes, we show the influence of the components by plotting the performance gap against the ablated models.
% As shown in Figure~\ref{fig:perforamnce_gap}, the medical knowledge contributed the most to the performance loss when applying the model trained on the CoNLL dataset and tested on the medical dataset, which is because such medical knowledge is highly related to the particular in-domain data.
% \textcolor{red}{what about other components?}
% As a summary, the cross-domain results suggest that our model has the potential to be applied to in particular domains when appropriate KGs are provided.

% \section{Case Study}

\section{Case Study}

To better illustrate the effectiveness of incorporating different knowledge in this task,
%into pronoun coreference resolution,
two examples are provided for the case study in Table~\ref{tab:case_study}.
In example A, our model correctly predicts that `\textit{it}' refers to the `\textit{magazine}' rather than the `\textit{room}', because we successfully retrieve the knowledge that compared with the `\textit{room}', the `\textit{magazine}' is more likely to be the object of \textit{drop}.
In example B, even though the distance between `erythema' and `This' is relatively far\footnote{We omit the intermediate part of the long sentence in the table for a clear presentation.}, our model is able to determine the coreference relation between them because it successfully finds out that `erythema' is a kind of disease, while a lot of diseases appear as the context of `be treated' in the training data.

\section{Related Work}

Detecting mention spans in linguistic expressions and identifying coreference relations among them is a core task, namely, coreference resolution, for natural language understanding.
Mention detection and coreference prediction are the two major focuses of the task as listed in \citet{DBLP:conf/emnlp/LeeHLZ17}.
Compared to general coreference problem,
pronoun coreference resolution has its unique challenge since
%as candidate noun phrases are already provided in the pronoun coreference task and
pronouns themselves have weak semantics meanings, which make it the most challenging sub-task in general coreference resolution.
To address the unique difficulty brought by pronouns, we thus focus on resolving pronoun coreferences in this paper.

% \begin{figure}[t]
%     \centering
%     \includegraphics[width=\linewidth]{image/Perforamnce_gap.png}
%     \caption{Performance loss of applying the model trained with a different domain data.}
%     \label{fig:perforamnce_gap}
% \end{figure}

\begin{table}[t]
    \centering
    \small
    \begin{tabular}{l|p{2.2cm}|p{2.6cm}}
    \toprule
         & Example A & Example B \\
    \midrule
        Sentence & 
        % The father takes the phone away from his \underline{\textcolor{blue}{son}} such that \textbf{\textcolor{red}{he}} can study. 
        He walks into the room with one \underline{\textcolor{blue}{magazine}} and drops \textbf{\textcolor{red}{it}} on the couch.
        & 
        ... A small area of \underline{\textcolor{blue}{erythema}} around his arm ... \textbf{\textcolor{red}{This}} will be treated empirically.\\
        % \midrule
        %  Pronoun & it & it \\
        %  \midrule
        %  Answer &  son & the therapy \\
         \midrule
         Prediction & magazine & erythema \\
         \midrule
        Knowledge & (`magazine', \textit{dobj},~~ `drop') & (`erythema', \textit{IsA}, `disease')\\
        
         \bottomrule
    \end{tabular}
    \caption{The case study on two examples from the test data, i.e., A: from the CoNLL and B: from the i2b2.
    Pronouns and correct mentions are marked by red bold and blue underline font respectively. Knowledge triplets used for them are listed in the bottom row.}
    \label{tab:case_study}
\end{table}

% Traditional approach. importance of knowledge

Resolving pronoun coreference relations often requires the support of manually crafted knowledge \cite{rahman2011coreference, emami2018hard}, especially for particular domains such as medicine \cite{uzuner2012evaluating} and biology \cite{cohen2017coreference}.
Previous studies on pronoun coreference resolution incorporated external knowledge including human defined rules \cite{hobbs1978resolving, ng2005supervised}, e.g., number/gender requirement of different pronouns, domain-specific knowledge such as medical~\cite{jindal2013end} or biological~\cite{trieu2018investigating} ones, and world knowledge \cite{rahman2011coreference}, such as selectional preference \cite{wilks1975preferential}.
%have been proved to be helpful for pronoun coreference resolution.
%
Later, end-to-end solutions \cite{DBLP:conf/emnlp/LeeHLZ17,lee2018higher} were proposed to 
%Recently, with the development of deep learning, \citet{DBLP:conf/emnlp/LeeHLZ17} proposed an end-to-end model that
learn contextual information and solve coreferences synchronously with neural networks, e.g., LSTM.
Their results proved that such knowledge is helpful when appropriately used for coreference resolution.
%when the context is properly encoded.
However, external knowledge is often omitted in their models.
Consider that 
%The aforementioned
context and external knowledge have their own advantages: the contextual information covering diverse text expressions that are difficult to be predefined while the external knowledge being usually more precisely constructed and able to provide extra information beyond the training data,
one could benefit from both sides for this task.
Different from previous studies,
we provide a generic solution to resolving pronoun coreference with the support of knowledge graphs based on contextual modeling,
%To the best of our knowledge, this is the first attempt that 
where deep learning models are adopted in our work to incorporate knowledge into pronoun coreference resolution and achieve remarkably good results.

\section{Conclusion}

In this paper, we explore how to build a knowledge-aware pronoun coreference resolution model, which is able to leverage different external knowledge for this task.
The proposed model is an attempt of the general solution
%To the proposed a general solution 
of incorporating knowledge (in the form of KG) into the deep learning based pronoun coreference model, rather than using knowledge as features or rules in a dedicated manner.
As a result, any knowledge resource presented
%we design our model to directly use the knowledge
in the format of triplets, the most widely used entry format for KG, can be consumed in our model with 
a proposed attention module.
%which is the most knowledge storage format of all modern knowledge graphs.
Experimental results on two different corpora from two domains demonstrate the superiority of the proposed model to all baselines.
Moreover, as our model learns to use knowledge rather than just fitting the training data, our model achieves much better and more robust performance than state-of-the-art models in the cross-domain scenario.
% As knowledge graph is not only useful for the pronoun coreference task, in the future, we plan to explore whether the proposed model can be helpful for other tasks.

\section*{Acknowledgements}
This paper was partially supported by the Early Career Scheme (ECS, No.26206717) from Research Grants Council in Hong Kong and Tencent AI Lab Rhino-Bird Focused Research Program. In addition, Hongming Zhang has been supported by the Hong Kong Ph.D. Fellowship.
We also thank 
% Intel Corporation for supporting our deep learning related research,
%We also thank
% and 
the anonymous reviewers for their valuable comments and suggestions that help improving the quality of this paper.

% \clearpage
\bibliography{pronoun-coreference-kg}
\bibliographystyle{acl_natbib}

\end{document}